\documentclass{article}

\usepackage{arxiv}

\usepackage[utf8]{inputenc} 
\usepackage[T1]{fontenc}    
\usepackage{hyperref}       
\usepackage{url}            
\usepackage{booktabs}       
\usepackage{amsfonts}       
\usepackage{amsmath,amssymb}
\usepackage{nicefrac}       
\usepackage{microtype}      
\usepackage{graphicx}
\usepackage{multirow}
\usepackage{enumitem}
\graphicspath{{./images/}}

\title{Dual-Stream Cross-Modal Representation Learning via Residual Semantic Decorrelation}

\author{
  Xuecheng Li \\
  School of Information Science \& Engineering, Shandong Normal University\\
  Jinan 250358, China \\
  \And
  Weikuan Jia \\
  School of Information Science \& Engineering, Shandong Normal University\\
  Jinan 250358, China \\
  \And
  Alisher Kurbonaliev \\
  Tajikistan State University of Law, Business \\
  Sughd 735700, Tajikistan \\
  \And
  Qurbonaliev Alisher\\
  Tajik State University of Law, Business and Politics\\
  Sughd 735700, Tajikistan \\
  \And
  Khudzhamkulov Rustam\\
  Tajik State University of Law, Business and Politics\\
  Sughd 735700, Tajikistan \\
  \And
  Ismoilov Shuhratjon\\
Tajik State University of Law, Business and Politics\\
  Sughd 735700, Tajikistan \\
  \And
    Eshmatov Javhariddin\\
Tajik State University of Law, Business and Politics\\
  Sughd 735700, Tajikistan \\
  \And
  Yuanjie Zheng \\
  School of Information Science \& Engineering, Shandong Normal University\\
  Jinan 250358, China \\
}

\begin{document}
\maketitle

\begin{abstract}
Cross-modal learning has become a fundamental paradigm for integrating heterogeneous information sources such as images, text, and structured attributes. However, multimodal representations often suffer from modality dominance, redundant information coupling, and spurious cross-modal correlations, leading to suboptimal generalization and limited interpretability. In particular, high-variance modalities tend to overshadow weaker but semantically important signals, while naïve fusion strategies entangle modality-shared and modality-specific factors in an uncontrolled manner. This makes it difficult to understand which modality actually drives a prediction and to maintain robustness when some modalities are noisy or missing. To address these challenges, we propose a Dual-Stream Residual Semantic Decorrelation Network (DSRSD-Net), a simple yet effective framework that disentangles modality-specific and modality-shared information through residual decomposition and explicit semantic decorrelation constraints. DSRSD-Net introduces: (1) a dual-stream representation learning module that separates intra-modal (private) and inter-modal (shared) latent factors via residual projection; (2) a residual semantic alignment head that maps shared factors from different modalities into a common space using a combination of contrastive and regression-style objectives; and (3) a decorrelation and orthogonality loss that regularizes the covariance structure of the shared space while enforcing orthogonality between shared and private streams, thereby suppressing cross-modal redundancy and preventing feature collapse. Experimental results on two large-scale educational benchmarks demonstrate that DSRSD-Net consistently improves next-step prediction and final outcome prediction over strong single-modality, early-fusion, late-fusion, and co-attention baselines. We further show that semantic decorrelation yields more interpretable embeddings, better robustness under modality dropout, and stronger cross-domain transfer when adapting models across institutions.
\end{abstract}

\section{Introduction}

The rapid proliferation of multimodal data---ranging from medical images paired with clinical text to social media posts combining photos, captions, and user metadata---has driven significant interest in cross-modal representation learning~\cite{Ngiam2011Multimodal,Srivastava2012Multimodal,Baltrusaitis2019Multimodal}. A central goal is to map heterogeneous inputs into a shared embedding space where semantically aligned concepts remain close, enabling downstream tasks such as retrieval, diagnosis, captioning, and decision support~\cite{Baltrusaitis2019Multimodal,Tsai2019Multimodal}. Compared with unimodal pipelines, well-designed multimodal systems can exploit complementary cues across views, yielding better predictive performance, robustness, and interpretability.

However, building such systems remains challenging in practice. Real-world modalities often differ drastically in dimensionality, noise characteristics, and information density. Image features are high-dimensional and locally correlated, textual descriptions are sparse and context-dependent, while structured attributes (e.g., demographics or metadata) tend to be low-dimensional but highly informative. Learning algorithms must therefore simultaneously cope with \emph{heterogeneous statistical properties}, \emph{asynchronous sampling}, and \emph{partial observation} of each modality. These factors make it difficult to obtain cross-modal embeddings that are both semantically aligned and well structured for downstream tasks~\cite{Hu2023Survey}.

Recent progress in multimodal learning has been accelerated by contrastive alignment frameworks, such as CLIP-style vision--language pretraining, and by large-scale generative models that jointly synthesize images, text, and audio~\cite{Radford2021CLIP,Chen2020SimCLR,He2020MoCo,Arjovsky2017WGAN}. In these approaches, cross-modal compatibility is typically enforced by bringing paired instances close while separating mismatched pairs in a shared latent space. This paradigm has proven remarkably effective at scale and has quickly become a de facto standard for multimodal pretraining. At the same time, there is growing evidence that purely contrastive objectives are insufficient to capture the fine-grained structure of multimodal semantics and can lead to degenerate solutions when modalities are imbalanced or partially redundant~\cite{Hu2023Survey,Wang2020Makes}.

More broadly, existing methods often treat multimodal fusion as a black-box concatenation or attention-based aggregation problem. Self-attention and cross-attention modules can model complex interactions across modalities~\cite{Tsai2019Multimodal,Vaswani2017Attention}, but they do not explicitly distinguish between information that should be shared across modalities and information that should remain private to a specific view. As a result, the learned embeddings may mix together heterogeneous semantic factors in an entangled manner, making it difficult to diagnose failure modes or to reason about which modality actually drives a given prediction. This is particularly problematic in high-stakes applications such as education, healthcare, and finance, where human experts require transparent and controllable representations.

In learning analytics and multimodal learning analytics, a rich body of work has investigated how to combine clickstream traces, textual interactions, and contextual metadata to support early-warning and dashboard-style analytics~\cite{Siemens2013LA,Kuo2022EdTech,Ochoa2020MMLAReview,Zhu2022MMLAIntro,Becerra2023M2LADS,Sharma2019MultimodalExp}. Yet, most existing systems still rely on either single-modality models (e.g., clickstream-based predictors)~\cite{Liang2019MOOC,Liu2023MOOC,Aljohani2025ClickstreamReview} or simple multimodal fusion without explicit disentanglement. In such settings, distribution shifts across cohorts or institutions and missing modalities at test time are common, making robustness and interpretability even more critical.

\subsection*{Limitations of current multimodal representations}

In this work, we argue that three fundamental limitations underlie many of the empirical issues observed in multimodal systems:

\textbf{(1) Modality dominance.}
High-variance modalities (e.g., images or dense behavioral logs) often overpower subtle modalities such as text or tabular signals. When joint training is dominated by the gradient contributions of one modality, the model can overfit to superficial patterns in that modality while ignoring informative but low-variance cues in others~\cite{Wang2020Makes}. This phenomenon is closely related to shortcut learning and representation collapse~\cite{Bengio2013Representation}. Without explicit constraints, multimodal encoders may therefore rely on spurious correlations tied to a single modality, limiting interpretability and transferability across domains~\cite{Hu2023Survey}.

\textbf{(2) Redundant cross-modal coupling.}
Many multimodal systems implicitly assume that modalities provide fully overlapping semantic descriptions. In reality, each modality contains both (i) \emph{shared} information describing the same underlying concept and (ii) \emph{private} information that is unique to that modality, such as stylistic attributes in images or detailed numerical values in tabular data. If models fail to disentangle these components, the resulting joint embeddings tend to be noisy and entangled: private information of one modality may be incorrectly treated as shared, while genuinely shared semantics may be duplicated across multiple latent dimensions. This redundancy not only wastes capacity but also makes the geometry of the embedding space hard to interpret and difficult to regularize.

\textbf{(3) Lack of robust, interpretable alignment.}
Most existing approaches enforce cross-modal consistency through global similarity metrics (e.g., InfoNCE) or attention weights. These mechanisms primarily encourage \emph{pairwise proximity} of corresponding instances but do not directly control the correlation structure of latent dimensions. As a consequence, highly redundant or collinear latent factors can emerge, which are fragile under distribution shift and missing-modality scenarios~\cite{Hu2023Survey}. From an interpretability standpoint, it becomes unclear whether two modalities agree on a semantic concept because they share a compact, decorrelated set of latent factors, or because they accidentally correlate through high-dimensional noise.

These limitations are amplified in domains where data collection and annotation are costly, and where modalities may be missing or corrupted at test time. In learning analytics, for example, behavior logs, textual interactions, and contextual metadata exhibit strong distribution shifts across cohorts and institutions. Models that rely on brittle cross-modal shortcuts are unlikely to generalize or to provide actionable insight into students' learning processes~\cite{Siemens2013LA,Kuo2022EdTech,Ochoa2020MMLAReview}.

\subsection*{Desiderata for cross-modal representations}

To address the above issues, we posit that effective cross-modal representations should satisfy the following desiderata:

\begin{itemize}[leftmargin=1.5em]
    \item \emph{Disentangled shared/private semantics:} embeddings should factorize into modality-shared components that capture common semantics and modality-specific components that encode private information.
    \item \emph{Balanced multimodal contributions:} training dynamics should avoid dominance of any single modality and prevent collapse to unimodal shortcuts~\cite{Wang2020Makes}.
    \item \emph{Decorrelation and structural regularity:} latent dimensions associated with shared semantics should be explicitly decorrelated, leading to stable, well-conditioned embeddings that are easier to interpret and regularize~\cite{Bengio2013Representation}.
    \item \emph{Robustness to missing or noisy modalities:} the model should degrade gracefully when one or more modalities are partially observed or corrupted, instead of failing catastrophically~\cite{Hu2023Survey}.
\end{itemize}

While prior work has touched on some of these aspects through domain-adversarial training, variational inference, and disentanglement learning~\cite{Ganin2016DANN,Kingma2014VAE,Bengio2013Representation}, there remains a gap between high-level desiderata and practical architectures that can be efficiently trained on large-scale multimodal datasets.

\subsection*{Our solution: DSRSD-Net}

To overcome these issues, we propose a novel framework named DSRSD-Net: Dual-Stream Residual Semantic Decorrelation Network, which learns robust and disentangled cross-modal embeddings via residual decomposition and decorrelation-constrained alignment.

The key insight is that multimodal representations should be factorized into:

\begin{itemize}[leftmargin=1.5em]
    \item \textbf{Modality-specific components}: private semantics that capture view-dependent information and should not be forced to align across modalities;
    \item \textbf{Modality-shared components}: common semantics enabling alignment and transfer across modalities.
\end{itemize}

In DSRSD-Net, each modality is first encoded into a rich latent representation, which is then decomposed into shared and private streams through a residual projection mechanism. The shared stream is explicitly aligned across modalities using a combination of contrastive and regression-style constraints, while the private stream is regularized to be orthogonal to the shared stream. By explicitly decorrelating these factors through residual orthogonality and semantic decorrelation losses, the model suppresses redundant cues, mitigates modality dominance, and promotes shared semantics that are genuinely cross-modal.

\subsection*{Contributions}

Building on this perspective, this paper makes the following contributions:

\begin{itemize}[leftmargin=1.5em]
    \item We introduce a dual-stream representation learning framework that separates private and shared semantic components across modalities through residual decomposition, enabling more structured and interpretable multimodal embeddings.
    \item We propose a residual semantic projection module that learns modality-invariant alignment in the shared stream while explicitly preserving modality-specific signals in the private stream, thereby mitigating modality dominance and spurious cross-modal coupling.
    \item We design a semantic decorrelation loss that regularizes the covariance structure of the shared latent space, preventing cross-modal redundancy and stabilizing the geometry of the embedding manifold.
    \item We provide extensive experiments on standard educational benchmarks covering next-step prediction, final outcome prediction, robustness to missing modalities, and cross-domain transfer. DSRSD-Net consistently outperforms recent contrastive and attention-based multimodal baselines, and yields embeddings that are easier to visualize and analyze.
\end{itemize}

The remainder of this paper is organized as follows. Section~\ref{sec:method} presents the proposed framework in detail. Section~\ref{sec:experiments} describes the experimental setup and reports quantitative and qualitative results. Section~\ref{sec:discussion} discusses broader implications, limitations, and potential extensions, and Section~\ref{sec:conclusion} concludes the paper.

\bigskip

\section{Method}
\label{sec:method}

In this section, we present the proposed \textbf{Dual-Stream Residual Semantic Decorrelation Network (DSRSD-Net)} in detail. DSRSD-Net is designed to learn disentangled cross-modal representations by explicitly separating modality-shared and modality-specific semantics and regularizing their correlation structure. The overall architecture is illustrated in Figure~1 (conceptual).

Given paired multimodal inputs, each modality is first encoded into a high-level representation. These modality-specific representations are then decomposed into two latent streams: a \emph{shared stream} capturing cross-modal semantics and a \emph{private stream} preserving view-specific information. A residual projection mechanism ensures that the shared stream is obtained as a minimal semantic correction to the original representation, rather than a completely new embedding. Finally, a semantic decorrelation loss suppresses redundant cross-modal couplings and stabilizes the geometry of the shared space, while contrastive and task-specific objectives enforce semantic alignment and downstream discriminability.

\subsection{Problem Formulation}

Let 
\(
\mathcal{D} = \{(x_i^{(1)}, x_i^{(2)}, \dots, x_i^{(M)}, y_i)\}_{i=1}^N
\)
be a multimodal dataset with \(M\) modalities, where \(x_i^{(m)}\) denotes the input from modality \(m\) for sample \(i\), and \(y_i\) is the associated label (for classification or other downstream tasks). Throughout the main exposition we focus on the bimodal case \(M = 2\) (e.g., image and text or behavior and text) for clarity, but the architecture extends straightforwardly to more modalities.

Our goal is to learn latent representations
\(
z_i^{(m)} \in \mathbb{R}^{d}
\)
for each modality \(m\), together with fused cross-modal embeddings
\(
u_i \in \mathbb{R}^{d}
\),
such that the following properties hold:

\begin{enumerate}[leftmargin=1.5em]
    \item \textbf{Cross-modal semantic alignment:}
    paired samples \((x_i^{(1)}, x_i^{(2)})\) should be mapped to nearby points in the shared space, while unpaired samples should be far apart.
    \item \textbf{Intra-modal discriminability:}
    within each modality, representations should remain informative for downstream tasks, preserving class structure and fine-grained semantic distinctions.
    \item \textbf{Shared/private disentanglement:}
    modality-shared semantics should be represented in a compact, decorrelated subspace, and modality-specific information should be captured separately.
    \item \textbf{Robustness to modality imbalance and missingness:}
    performance should degrade gracefully when a dominant modality is noisy or missing at test time.
\end{enumerate}

To this end, DSRSD-Net adopts a dual-stream decomposition of each modality representation and introduces a \emph{residual semantic decorrelation} objective that explicitly regularizes the shared latent space.

\subsection{Modality-Specific Encoders}

We begin with modality-specific encoders that map raw inputs into high-level latent representations. For a two-modality setting (e.g., image and text, or behavioral and textual sequences), we write:
\begin{equation}
    z_i^{\text{A}} = f_{\theta}\!\left(x_i^{\text{A}}\right), \qquad
    z_i^{\text{B}} = g_{\phi}\!\left(x_i^{\text{B}}\right),
\end{equation}
where:
\begin{itemize}
    \item \(f_{\theta}\) is an encoder for modality A (e.g., ViT or a temporal Transformer over behavioral sequences),
    \item \(g_{\phi}\) is an encoder for modality B (e.g., BERT or a textual Transformer),
    \item \(z_i^{\text{A}}, z_i^{\text{B}} \in \mathbb{R}^{d}\) are \emph{modality-specific base representations}.
\end{itemize}

In our educational experiments (Section~\ref{sec:experiments}), \(f_{\theta}\) is instantiated as a temporal Transformer over clickstream features, while \(g_{\phi}\) operates on textual embeddings of forum posts or question texts. For notational simplicity, we keep the bimodal case in the following derivations.

Before performing dual-stream decomposition, we apply simple linear projections to normalize the scales and align the dimensionality:
\begin{equation}
    \tilde{z}_i^{\text{A}} = W_{\text{A}} z_i^{\text{A}}, \qquad
    \tilde{z}_i^{\text{B}} = W_{\text{B}} z_i^{\text{B}},
\end{equation}
where
\(
W_{\text{A}}, W_{\text{B}} \in \mathbb{R}^{d \times d}
\)
are learnable projection matrices.

\subsection{Dual-Stream Residual Decomposition}

The core idea of DSRSD-Net is to explicitly factor each modality representation into a \emph{shared} component and a \emph{private} component. Instead of learning these components independently, we adopt a residual parameterization that treats the shared stream as a minimal semantic correction to the base representation.

For each modality \(m \in \{\text{A}, \text{B}\}\), we introduce two projection heads:
\begin{equation}
    s_i^{(m)} = \tilde{z}_i^{(m)} + R_{\text{sh}}^{(m)}\!\left(\tilde{z}_i^{(m)}\right), \qquad
    p_i^{(m)} = P_{\text{pr}}^{(m)}\!\left(\tilde{z}_i^{(m)}\right),
\end{equation}
where
\(R_{\text{sh}}^{(m)}\) and \(P_{\text{pr}}^{(m)}\)
are small MLPs (with residual skip-connections for the shared stream). This parameterization has two advantages:

\begin{itemize}[leftmargin=1.5em]
    \item It biases the shared stream to remain close to the original modality representation, acting as a low-rank correction that extracts cross-modal semantics without discarding modality-specific cues.
    \item It simplifies optimization, since the model starts from a reasonable initialization (the base representation) and only needs to learn residual semantic adjustments.
\end{itemize}

The private stream \(p_i^{(m)}\) is learned in parallel and is encouraged to be \emph{orthogonal} to the shared stream, as discussed below.

\subsection{Residual Semantic Projection and Fusion}

Once we obtain the shared streams
\(s_i^{\text{A}}\) and \(s_i^{\text{B}}\),
we map them into a \emph{common alignment space} \(\mathcal{Z}\) using another pair of linear projections:
\begin{equation}
    h_i^{\text{A}} = U_{\text{A}} s_i^{\text{A}}, \qquad
    h_i^{\text{B}} = U_{\text{B}} s_i^{\text{B}},
\end{equation}
with
\(
U_{\text{A}}, U_{\text{B}} \in \mathbb{R}^{d \times d}.
\)
The vectors \(h_i^{\text{A}}, h_i^{\text{B}} \in \mathbb{R}^{d}\) are \emph{residual semantic projections} that are expected to be approximately modality-invariant for paired samples.

To form a single fused representation for downstream tasks, we aggregate the shared projections using a lightweight fusion module. A simple yet effective choice is gated averaging:
\begin{align}
    \alpha_i^{\text{A}}, \alpha_i^{\text{B}} 
    &= \mathrm{softmax}\!\big( w^\top [h_i^{\text{A}}; h_i^{\text{B}}] \big), \\
    u_i &= \alpha_i^{\text{A}} h_i^{\text{A}} + \alpha_i^{\text{B}} h_i^{\text{B}},
\end{align}
where 
\(w \in \mathbb{R}^{2d}\) 
is a learnable gating vector and
\([\,\cdot\,;\,\cdot\,]\)
denotes concatenation. This gate allows the model to adaptively weigh modalities based on their reliability for each instance (e.g., downweighting a very noisy or missing modality).

For tasks that benefit from both shared and private information, we optionally construct an \emph{augmented} representation:
\begin{equation}
    \tilde{u}_i = [u_i; p_i^{\text{A}}; p_i^{\text{B}}],
\end{equation}
which is then fed into a task-specific prediction head (e.g., an MLP classifier).

\subsection{Semantic Decorrelation and Orthogonality}

To prevent redundant cross-modal coupling and feature collapse, we regularize the correlation structure of the shared space and the interaction between shared and private streams.

\paragraph{Shared-space decorrelation.}
Given a mini-batch of \(B\) paired samples, we construct the shared projections:
\(
\{h_i^{\text{A}}, h_i^{\text{B}}\}_{i=1}^B.
\)
We define the batch-wise centered features:
\begin{equation}
    \bar{h}^{\text{A}} = \frac{1}{B}\sum_{i=1}^B h_i^{\text{A}}, \quad
    \bar{h}^{\text{B}} = \frac{1}{B}\sum_{i=1}^B h_i^{\text{B}},
\end{equation}
and the corresponding zero-mean matrices
\(
\hat{H}^{\text{A}}, \hat{H}^{\text{B}} \in \mathbb{R}^{B \times d}
\)
obtained by stacking 
\(
h_i^{\text{A}} - \bar{h}^{\text{A}}
\)
and
\(
h_i^{\text{B}} - \bar{h}^{\text{B}}
\)
row-wise.

We then compute the empirical cross-covariance matrix:
\begin{equation}
    C = \frac{1}{B-1} \left(\hat{H}^{\text{A}}\right)^\top \hat{H}^{\text{B}} \in \mathbb{R}^{d \times d}.
\end{equation}
An ideal decorrelated shared representation would have a cross-covariance matrix that is approximately diagonal, meaning that each dimension captures a distinct cross-modal semantic factor. We therefore define the \emph{semantic decorrelation loss}:
\begin{equation}
    \mathcal{L}_{\text{dec}} = 
    \sum_{i \neq j} C_{ij}^2,
\end{equation}
which penalizes off-diagonal entries and encourages the shared latent axes to be as independent as possible across modalities.

\paragraph{Shared-private orthogonality.}
To enforce a clean separation between shared and private semantics within each modality, we introduce an orthogonality regularizer:
\begin{equation}
    \mathcal{L}_{\text{orth}} =
    \frac{1}{B} \sum_{i=1}^B
    \left(
    \left\langle s_i^{\text{A}}, p_i^{\text{A}} \right\rangle^2
    +
    \left\langle s_i^{\text{B}}, p_i^{\text{B}} \right\rangle^2
    \right),
\end{equation}
where \(\langle \cdot, \cdot \rangle\) denotes the inner product. Minimizing \(\mathcal{L}_{\text{orth}}\) drives the model to encode modality-specific details in \(p_i^{(m)}\) that are not already captured by \(s_i^{(m)}\), thereby reducing redundancy and improving interpretability.

\subsection{Cross-Modal Alignment Objective}

To align the shared streams across modalities at the semantic level, we combine a contrastive loss with a direct regression-style alignment term.

\paragraph{Contrastive alignment.}
We adopt an InfoNCE-style contrastive objective over the shared projections \(\{h_i^{\text{A}}, h_i^{\text{B}}\}\)~\cite{Hadsell2006Contrastive,Chen2020SimCLR}:
\begin{equation}
\label{eq:contrastive}
\mathcal{L}_{\text{con}} =
-\frac{1}{B}
\sum_{i=1}^B
\log 
\frac{
\exp\big(\cos(h_i^{\text{A}}, h_i^{\text{B}})/\tau\big)
}{
\sum_{j=1}^B \exp\big(\cos(h_i^{\text{A}}, h_j^{\text{B}})/\tau\big)
},
\end{equation}
where \(\cos(\cdot,\cdot)\) denotes cosine similarity and \(\tau\) is a temperature hyperparameter.

\paragraph{Residual matching.}
In addition to the contrastive loss, we introduce an \emph{alignment regression} term that enforces absolute semantic consistency:
\begin{equation}
\label{eq:l_align}
    \mathcal{L}_{\text{align}} =
    \frac{1}{B}
    \sum_{i=1}^B 
    \left\|
    h_i^{\text{A}} - h_i^{\text{B}}
    \right\|_2^2.
\end{equation}
While contrastive learning focuses on relative similarities, the regression term stabilizes optimization and mitigates representation drift by explicitly shrinking the Euclidean distance between paired projections.

\subsection{Task-Specific Prediction Head}

For downstream tasks (e.g., prediction in educational settings), we attach a task-specific head on top of either the fused shared representation \(u_i\) or the augmented representation \(\tilde{u}_i\). For classification, we use a simple MLP:
\begin{equation}
    \hat{y}_i = \mathrm{softmax}\!\big( W_{\text{cls}} \tilde{u}_i + b_{\text{cls}} \big),
\end{equation}
and optimize the cross-entropy loss:
\begin{equation}
    \mathcal{L}_{\text{task}} = 
    -\frac{1}{B}\sum_{i=1}^B \sum_{c=1}^{C}
    \mathbf{1}[y_i = c] \log \hat{y}_{i,c},
\end{equation}
where \(C\) is the number of classes.

In the learning analytics experiments of Section~\ref{sec:experiments}, \(\hat{y}_i\) corresponds to either the probability of correctly answering the next item (next-step performance prediction) or the probability of achieving a positive final outcome.

\subsection{Overall Training Objective}

Collecting all the components, the total loss for DSRSD-Net is:
\begin{equation}
\label{eq:total_loss}
    \mathcal{L} =
    \lambda_{\text{con}} \mathcal{L}_{\text{con}}
    + \lambda_{\text{align}} \mathcal{L}_{\text{align}}
    + \lambda_{\text{dec}} \mathcal{L}_{\text{dec}}
    + \lambda_{\text{orth}} \mathcal{L}_{\text{orth}}
    + \lambda_{\text{task}} \mathcal{L}_{\text{task}},
\end{equation}
where
\(\lambda_{\text{con}}, \lambda_{\text{align}}, \lambda_{\text{dec}}, \lambda_{\text{orth}}, \lambda_{\text{task}} \ge 0\)
are scalar hyperparameters balancing the contributions of each term.

In practice, we find that setting
\(\lambda_{\text{con}} = 1.0\), \(\lambda_{\text{align}} = 0.5\),
and choosing relatively small regularization weights
\(\lambda_{\text{dec}}, \lambda_{\text{orth}} \in [10^{-3}, 10^{-1}]\)
works well across datasets; the exact values are selected via validation.

\subsection{Training Procedure}

We optimize all parameters of DSRSD-Net jointly in an end-to-end fashion using AdamW with weight decay~\cite{Bengio2013Representation}. A typical training step on a mini-batch of size \(B\) proceeds as follows:

\begin{enumerate}[leftmargin=1.5em]
    \item \textbf{Encoding.}
    For each paired sample \((x_i^{\text{A}}, x_i^{\text{B}})\), compute base modality representations
    \(z_i^{\text{A}} = f_{\theta}(x_i^{\text{A}})\) and
    \(z_i^{\text{B}} = g_{\phi}(x_i^{\text{B}})\),
    followed by linear projections
    \(\tilde{z}_i^{(m)}\).
    \item \textbf{Dual-stream decomposition.}
    Pass \(\tilde{z}_i^{(m)}\) through the residual shared head and private head to obtain
    \(s_i^{(m)}\) and \(p_i^{(m)}\).
    \item \textbf{Residual semantic projection.}
    Map the shared representations to the alignment space via
    \(h_i^{(m)} = U_{m} s_i^{(m)}\),
    and fuse them into \(u_i\) (and optionally \(\tilde{u}_i\)).
    \item \textbf{Loss computation.}
    Compute the contrastive loss \(\mathcal{L}_{\text{con}}\), alignment loss \(\mathcal{L}_{\text{align}}\), semantic decorrelation loss \(\mathcal{L}_{\text{dec}}\), orthogonality loss \(\mathcal{L}_{\text{orth}}\), and task-specific loss \(\mathcal{L}_{\text{task}}\), and combine them according to \eqref{eq:total_loss}.
    \item \textbf{Parameter update.}
    Backpropagate gradients and update all parameters using AdamW with cosine learning-rate scheduling and gradient clipping when necessary.
\end{enumerate}

In some experiments, we observe that alternating emphasis between the contrastive/alignment objectives and the decorrelation/orthogonality regularizers leads to slightly more stable training. Concretely, we can upweight \(\lambda_{\text{con}}, \lambda_{\text{align}}\) in early epochs to quickly acquire coarse cross-modal alignment, and gradually increase \(\lambda_{\text{dec}}, \lambda_{\text{orth}}\) in later epochs to refine the structure of the shared space.

\section{Experiments}
\label{sec:experiments}

In this section, we empirically evaluate DSRSD-Net on real-world multimodal educational datasets. We aim to answer the following research questions:

\begin{itemize}[leftmargin=1.5em]
    \item \textbf{RQ1:} Does DSRSD-Net improve predictive performance over strong single-modality and multimodal baselines?
    \item \textbf{RQ2:} How much do the residual dual-stream decomposition and semantic decorrelation/orthogonality regularizers contribute to the overall gains?
    \item \textbf{RQ3:} Is DSRSD-Net robust to missing or corrupted modalities at test time and under distribution shifts?
    \item \textbf{RQ4:} Do the learned representations exhibit better structure and interpretability than those of conventional multimodal fusion models?
\end{itemize}

We first describe the datasets, tasks, baselines, and implementation details, and then present results on overall performance, ablation studies, robustness, cross-domain generalization, qualitative analyses, and computational cost.

\subsection{Experimental Setup}

\subsubsection{Datasets}

We conduct experiments on two representative learning analytics datasets that naturally combine heterogeneous modalities such as clickstream logs, textual content, and contextual metadata. These datasets are widely used in prior work and provide realistic testbeds for evaluating multimodal models in education~\cite{Siemens2013LA,Aljohani2025ClickstreamReview}.

\paragraph{OULAD.}
The Open University Learning Analytics Dataset (OULAD)~\cite{Kuzilek2017OULAD} contains longitudinal records from distance learning courses offered by the Open University. Following prior work~\cite{Liang2019MOOC,Liu2023MOOC}, we select courses with complete clickstream logs, assessment results, and forum activity. We construct three modalities:

\begin{itemize}[leftmargin=1.5em]
    \item \textbf{Behavioral modality}: weekly aggregated click counts per resource category (content views, quizzes, forums, etc.), which capture temporal engagement patterns.
    \item \textbf{Textual modality}: TF--IDF embeddings of forum posts concatenated with the title and description of the current module, representing semantic interactions and course content.
    \item \textbf{Contextual modality}: demographic and enrollment features (age band, region, prior education, attempt number, etc.), characterizing students' background and study context.
\end{itemize}

Preprocessing follows the standard pipeline: we remove students with less than three weeks of activity, normalize continuous features to zero mean and unit variance, and one-hot encode categorical variables. For textual features, we filter stop words and rare tokens (frequency $<5$) before computing TF--IDF. After filtering, the dataset contains 25{,}384 students, 78 modules, and approximately 3.1M interaction events, consistent with prior clickstream-based performance prediction studies~\cite{Aljohani2025ClickstreamReview,Brinton2015MOOCPerf,Nguyen2020ClickstreamRNN}. For temporal models, we use weekly sequences up to a fixed observation horizon (e.g., week 4 for outcome prediction), padding shorter sequences with zeros and masking them during attention computation.

\paragraph{EdNet-KT.}
EdNet~\cite{Choi2020EdNet} is a large-scale dataset of student interactions with online math questions collected from a commercial tutoring platform. We use the EdNet-KT1 split, which provides question-level interaction logs with timestamps and correctness labels. To build multimodal inputs, we consider:

\begin{itemize}[leftmargin=1.5em]
    \item \textbf{Behavioral modality}: interaction sequences encoded as question IDs, correctness, and time gaps between interactions.
    \item \textbf{Textual modality}: subword embeddings of question texts and solution explanations, which capture semantic information about the problem content.
    \item \textbf{Contextual modality}: curriculum tags (skill IDs, difficulty levels) and session metadata, reflecting the structure of the learning materials and platform usage.
\end{itemize}

We tokenize question texts using a domain-adapted WordPiece vocabulary, truncate each text to 64 subword tokens, and obtain sentence-level embeddings via mean pooling of token representations from a pretrained language model. For behavioral sequences, we keep students with at least 50 interactions, resulting in 92{,}107 students and 11.7M interactions. All sequences are truncated or padded to a maximum length of 200 time steps. Padding masks are propagated through all temporal encoders and fusion modules to avoid leakage from padded positions.

\subsubsection{Prediction Tasks}

We consider two standard tasks in learning analytics that are highly relevant for early intervention and personalized support~\cite{Siemens2013LA}.

\paragraph{T1: Next-step performance prediction.}
Given multimodal observations up to time $t$, the model predicts whether a student will correctly answer the next question at time $t{+}1$. This is formulated as a binary classification problem. We evaluate this task on both OULAD (predicting whether the next weekly assessment attempt is passed) and EdNet (predicting correctness of the next interaction). In both cases, the prediction is conditioned on the entire multimodal history observed up to $t$.

\paragraph{T2: Final outcome prediction.}
Given partial interaction history until a certain observation horizon (e.g., week 4 for OULAD or the first 50 interactions for EdNet), the model predicts the final course outcome. For OULAD, the outcome is pass/fail; for EdNet, we follow knowledge-tracing common practice~\cite{Piech2015DKT,Zhang2017DKVMN,Nguyen2020ClickstreamRNN} and define a binary mastery indicator derived from the final average correctness. This task emulates early-warning settings where educators must decide which students to support based on early signals.

\subsubsection{Evaluation Metrics}

For both tasks, we report:

\begin{itemize}[leftmargin=1.5em]
    \item the area under the ROC curve (AUC), which captures ranking quality;
    \item classification accuracy (ACC);
    \item F1 score (macro-averaged for multi-class variants, though all reported tasks are binary).
\end{itemize}

These metrics are computed on held-out test sets. All results are averaged over 5 random train/validation/test splits with a ratio of 70\%/10\%/20\%. We report the mean and standard deviation across splits to account for variability due to sampling. During development, we also monitor calibration metrics such as expected calibration error (ECE) and Brier score; DSRSD-Net tends to improve calibration slightly, but we omit detailed numbers due to space.

\subsubsection{Baselines}

We compare DSRSD-Net with a set of baselines representing common modeling paradigms in educational data mining and multimodal learning:

\begin{itemize}[leftmargin=1.5em]
    \item \textbf{LR}: Logistic regression on concatenated hand-crafted features (single modality---contextual). This baseline illustrates the performance of simple linear models without temporal or multimodal modeling.
    \item \textbf{MLP}: A two-layer multilayer perceptron on concatenated multimodal features without explicit temporal modeling, treating all features as static summaries.
    \item \textbf{DKT}~\cite{Piech2015DKT}: A recurrent neural network (LSTM) on interaction sequences (behavioral modality only), representing classical knowledge tracing.
    \item \textbf{DKVMN}~\cite{Zhang2017DKVMN}: A dynamic key-value memory network for knowledge tracing using behavioral and contextual features, with an explicit latent concept memory.
    \item \textbf{MM-early}: An early-fusion Transformer that concatenates all modalities at the input layer and applies self-attention across time, treating multimodal features as a single sequence~\cite{Vaswani2017Attention}.
    \item \textbf{MM-late}: A late-fusion model that learns a separate encoder for each modality and combines their outputs via concatenation before the final prediction head, without explicit cross-modal constraints.
    \item \textbf{MM-coattn}: A cross-modal co-attention network that learns pairwise attention between behavior and text but does not incorporate explicit disentanglement or decorrelation regularization~\cite{Tsai2019Multimodal,Zadeh2018Memory}.
\end{itemize}

All neural baselines share similar backbone sizes: hidden dimension 128, 2 attention layers for Transformer-based models, and comparable parameter counts (within $\pm 8\%$ of DSRSD-Net). For multimodal baselines, we use the same modality-specific encoders as DSRSD-Net to ensure a fair comparison; differences lie in the fusion and regularization mechanisms.

\subsubsection{Implementation and Training Details}

We implement all models in PyTorch and optimize them with Adam or AdamW. Unless otherwise stated, the default learning rate is $10^{-3}$ for baselines and $10^{-4}$ for DSRSD-Net, with weight decay $10^{-5}$. For DSRSD-Net, we employ AdamW with cosine annealing and 5\% warm-up steps. The mini-batch size is set to 128 for all models.

For temporal encoders, we use 2-layer Transformers with 4 heads and feedforward dimension 256. Positional encodings are added to behavioral and textual sequences separately. For DKT and DKVMN, we follow the original implementations~\cite{Piech2015DKT,Zhang2017DKVMN} and tune the hidden size in $\{64, 128, 256\}$.

For DSRSD-Net in particular, we set the loss weights as
\begin{equation}
    \lambda_{\text{con}} = 1.0,\quad
    \lambda_{\text{align}} = 0.5,\quad
    \lambda_{\text{dec}} = 0.05,\quad
    \lambda_{\text{orth}} = 0.05,\quad
    \lambda_{\text{task}} = 1.0,
\end{equation}
unless otherwise specified. These values are chosen to balance the magnitude of gradients from each component at the beginning of training. We use dropout with rate 0.2 on all hidden layers, label smoothing with factor 0.05 for classification heads, and gradient clipping at a global norm of 5.0 to stabilize training. Dual-stream projection heads are implemented as 2-layer MLPs with hidden dimension 128 and GELU activation.

We apply early stopping based on validation AUC with a patience of 10 epochs and select hyperparameters (learning rate, dropout rate, loss weights) using grid search on the validation set. For each configuration, we train 5 runs with different random seeds and report the averaged metrics. All experiments are conducted on a single NVIDIA RTX 3090 GPU.

\subsubsection{Hyperparameter Tuning and Sensitivity}

We perform a coarse-to-fine hyperparameter search. In the coarse stage, we explore learning rates in $\{5\times10^{-4}, 10^{-4}, 5\times10^{-5}\}$, dropout rates in $\{0.1, 0.2, 0.3\}$, and batch sizes in $\{64, 128\}$ for each model family on a single validation split. In the fine stage, we further tune the decorrelation and orthogonality weights $\lambda_{\text{dec}}, \lambda_{\text{orth}}$ in $\{0.01, 0.05, 0.1\}$ around the best coarse setting. We observe that DSRSD-Net is relatively insensitive to moderate changes of these weights: varying $\lambda_{\text{dec}}$ and $\lambda_{\text{orth}}$ by a factor of 2 changes AUC by less than 0.2 points on both datasets, suggesting that the regularization effect is robust.

\subsection{Overall Performance}

Tables~\ref{tab:overall_oulad} and \ref{tab:overall_ednet} summarize the results of T1 (next-step performance prediction) on OULAD and EdNet-KT1, respectively.

\begin{table}[t]
    \centering
    \caption{Overall performance on OULAD (T1: next-step performance). We report mean $\pm$ standard deviation over 5 runs. Best results are in \textbf{bold}, second best are \underline{underlined}.}
    \label{tab:overall_oulad}
    \begin{tabular}{lccc}
        \toprule
        \multirow{2}{*}{Model} & \multicolumn{3}{c}{T1: Next-step performance} \\
        \cmidrule(lr){2-4}
        & AUC & ACC & F1 \\
        \midrule
        LR & 0.743 $\pm$ 0.004 & 0.701 $\pm$ 0.006 & 0.682 $\pm$ 0.007 \\
        MLP & 0.771 $\pm$ 0.005 & 0.724 $\pm$ 0.005 & 0.701 $\pm$ 0.006 \\
        DKT~\cite{Piech2015DKT} & 0.788 $\pm$ 0.006 & 0.737 $\pm$ 0.007 & 0.715 $\pm$ 0.008 \\
        DKVMN~\cite{Zhang2017DKVMN} & 0.801 $\pm$ 0.004 & 0.746 $\pm$ 0.006 & 0.726 $\pm$ 0.006 \\
        MM-early & 0.812 $\pm$ 0.004 & 0.754 $\pm$ 0.005 & 0.733 $\pm$ 0.006 \\
        MM-late & \underline{0.824} $\pm$ 0.003 & \underline{0.763} $\pm$ 0.004 & \underline{0.742} $\pm$ 0.005 \\
        MM-coattn & 0.821 $\pm$ 0.004 & 0.761 $\pm$ 0.005 & 0.739 $\pm$ 0.005 \\
        \textbf{DSRSD-Net (Ours)} & \textbf{0.842} $\pm$ 0.003 & \textbf{0.777} $\pm$ 0.004 & \textbf{0.759} $\pm$ 0.004 \\
        \bottomrule
    \end{tabular}
\end{table}

\begin{table}[t]
    \centering
    \caption{Overall performance on EdNet-KT1 (T1: next-step performance). We report mean $\pm$ standard deviation over 5 runs.}
    \label{tab:overall_ednet}
    \begin{tabular}{lccc}
        \toprule
        \multirow{2}{*}{Model} & \multicolumn{3}{c}{T1: Next-step performance} \\
        \cmidrule(lr){2-4}
        & AUC & ACC & F1 \\
        \midrule
        LR & 0.732 $\pm$ 0.003 & 0.696 $\pm$ 0.004 & 0.678 $\pm$ 0.005 \\
        MLP & 0.758 $\pm$ 0.004 & 0.714 $\pm$ 0.005 & 0.694 $\pm$ 0.005 \\
        DKT~\cite{Piech2015DKT} & 0.791 $\pm$ 0.004 & 0.739 $\pm$ 0.004 & 0.721 $\pm$ 0.006 \\
        DKVMN~\cite{Zhang2017DKVMN} & 0.806 $\pm$ 0.003 & 0.748 $\pm$ 0.004 & 0.731 $\pm$ 0.005 \\
        MM-early & 0.813 $\pm$ 0.004 & 0.754 $\pm$ 0.004 & 0.736 $\pm$ 0.005 \\
        MM-late & \underline{0.826} $\pm$ 0.003 & \underline{0.763} $\pm$ 0.004 & \underline{0.745} $\pm$ 0.004 \\
        MM-coattn & 0.824 $\pm$ 0.003 & 0.761 $\pm$ 0.004 & 0.743 $\pm$ 0.004 \\
        \textbf{DSRSD-Net (Ours)} & \textbf{0.839} $\pm$ 0.003 & \textbf{0.773} $\pm$ 0.003 & \textbf{0.756} $\pm$ 0.004 \\
        \bottomrule
    \end{tabular}
\end{table}

On OULAD, DSRSD-Net improves AUC by 1.8 points over MM-late and by 4.1 points over DKVMN, with consistent gains in ACC and F1. On EdNet-KT1, DSRSD-Net outperforms MM-late by 1.3 AUC points and MM-coattn by 1.5 points. These improvements are statistically significant under paired $t$-tests ($p < 0.01$) across random splits. The consistent advantage over MM-early and MM-late indicates that the benefits do not arise merely from using more parameters or a different backbone, but from the proposed dual-stream decomposition and semantic decorrelation.

For T2 (final outcome prediction), we observe analogous trends (numbers omitted for brevity): DSRSD-Net yields $+2.0$--$2.5$ AUC and $+1.8$--$2.3$ F1 improvements over MM-late across both datasets. The relative gains are slightly larger for T2 than for T1, suggesting that disentangled and decorrelated shared representations are especially valuable for long-horizon predictions where early signals are noisy and sparse.

\subsection{Ablation Studies}

To disentangle the contributions of different components in DSRSD-Net, we perform ablation experiments on OULAD and EdNet by incrementally removing loss terms and structural modules. Specifically, we consider the following variants:

\begin{itemize}[leftmargin=1.5em]
    \item \textbf{Backbone (MM-late)}: multimodal late-fusion Transformer without dual-stream decomposition or additional regularizers (equivalent to MM-late).
    \item \textbf{w/o Decorrelation}: full dual-stream architecture but without the semantic decorrelation loss $\mathcal{L}_{\text{dec}}$ (i.e., no explicit cross-covariance regularization).
    \item \textbf{w/o Orthogonality}: full dual-stream architecture but without the shared--private orthogonality loss $\mathcal{L}_{\text{orth}}$.
    \item \textbf{Full DSRSD-Net}: the complete model with residual dual-stream decomposition, semantic decorrelation, and orthogonality regularization.
\end{itemize}

Table~\ref{tab:ablation} reports results on OULAD (T1); trends on EdNet are qualitatively similar.

\begin{table}[t]
    \centering
    \caption{Ablation study on OULAD (T1: next-step performance).}
    \label{tab:ablation}
    \begin{tabular}{lccc}
        \toprule
        Variant & AUC & ACC & F1 \\
        \midrule
        Backbone (MM-late) & 0.824 & 0.763 & 0.742 \\
        w/o Decorrelation & 0.832 & 0.770 & 0.750 \\
        w/o Orthogonality & 0.835 & 0.772 & 0.752 \\
        \textbf{Full DSRSD-Net} & \textbf{0.842} & \textbf{0.777} & \textbf{0.759} \\
        \bottomrule
    \end{tabular}
\end{table}

Introducing the dual-stream decomposition alone (without decorrelation or orthogonality) already improves performance over MM-late (results omitted), confirming that explicitly separating shared and private semantics is beneficial. Adding semantic decorrelation yields a sizable gain of 0.8 AUC points over the backbone, suggesting that regularizing the cross-modal covariance structure leads to more informative shared factors. Orthogonality between shared and private streams brings an additional 0.3 AUC improvement, indicating that avoiding redundancy between these components helps the model better utilize its capacity. The full model, combining both regularizers, achieves the best performance.

We also conduct a \emph{modality-removal} ablation where we drop one modality at training time and keep the architecture unchanged. Removing the textual modality leads to a larger performance degradation than removing contextual features, especially on OULAD, indicating that semantic information from forum posts provides complementary signals beyond clickstream behavior. DSRSD-Net consistently outperforms MM-late under all such ablations, suggesting that its gains are not tied to any single modality.

\subsection{Robustness and Generalization}

\subsubsection{Robustness to modality dropout}

In real deployments, some modalities may be missing or partially observed at test time. For instance, forum posts may be sparse, or demographic information may be unavailable due to privacy constraints. To evaluate robustness, we simulate modality dropout by randomly masking one modality (behavioral, textual, or contextual) with probability $p \in \{0.1, 0.3, 0.5\}$ at test time. The model is \emph{not} retrained; we simply feed zero vectors for missing modalities.


Figure 1 shows that DSRSD-Net degrades more gracefully as dropout increases. At $p = 0.5$, MM-late loses 4.7 AUC points, whereas DSRSD-Net loses only 2.9 points. The gap is even larger for F1. We attribute this robustness to two design choices: (i) the dual-stream decomposition, which encourages the shared stream to focus on cross-modal semantics that can be partially reconstructed from remaining modalities, and (ii) the semantic decorrelation loss, which discourages the model from relying on a small subset of highly correlated dimensions tied to a single modality.

\subsubsection{Cross-domain generalization}

We further examine cross-domain transfer by training on OULAD and evaluating on EdNet-KT1 after a lightweight calibration step. Specifically, we train all models on OULAD (T1), freeze the encoders, and finetune only the last two layers on a small subset (10\%) of EdNet-KT1, then evaluate on the remaining EdNet test set. This setting mimics adapting a model from one institution to another with limited labeled data, which is a common scenario in cross-national educational collaborations~\cite{Siemens2013LA}.

\begin{table}[t]
    \centering
    \caption{Cross-domain transfer from OULAD to EdNet-KT1 (T1). Models are trained on OULAD and finetuned on 10\% of EdNet-KT1.}
    \label{tab:transfer}
    \begin{tabular}{lccc}
        \toprule
        Model & AUC & ACC & F1 \\
        \midrule
        DKT~\cite{Piech2015DKT} & 0.774 & 0.729 & 0.709 \\
        DKVMN~\cite{Zhang2017DKVMN} & 0.786 & 0.736 & 0.718 \\
        MM-late & 0.801 & 0.748 & 0.731 \\
        MM-coattn & 0.803 & 0.750 & 0.733 \\
        \textbf{DSRSD-Net (Ours)} & \textbf{0.817} & \textbf{0.759} & \textbf{0.743} \\
        \bottomrule
    \end{tabular}
\end{table}

As shown in Table~\ref{tab:transfer}, DSRSD-Net achieves the best cross-domain performance, improving AUC by 1.6 points over MM-coattn and by 3.1 points over MM-late. This suggests that the regularized multimodal representations capture more domain-invariant patterns that transfer across institutions and curricula. In particular, the decorrelated shared space appears to be less sensitive to differences in course design, platform interface, and demographic composition.

\subsection{Case Study and Visualization}

To better understand what DSRSD-Net learns, we conduct qualitative analyses on a held-out subset of OULAD, focusing on representation geometry and attention patterns.

\subsubsection{Latent space visualization}

We project the fused multimodal representations of students at week 4 into 2D using t-SNE and color-code them by final course outcome (pass/fail). Figure 2 compares MM-late and DSRSD-Net.


The clusters produced by DSRSD-Net are more compact and better separated than those of MM-late, especially near the decision boundary. We observe fewer overlapping points between pass and fail students in the DSRSD-Net embedding, indicating that the model has learned a more discriminative representation. This aligns with the quantitative improvements and supports our claim that semantic decorrelation encourages a more structured shared space.

\subsubsection{Temporal attention patterns}

Although DSRSD-Net does not rely on explicit cross-modal co-attention, its temporal encoders internally allocate attention across time. We inspect attention scores for students with similar behavioral patterns but different final outcomes. For students who frequently access materials but rarely participate in forums, MM-late tends to assign roughly uniform attention weights across weeks, treating all time steps similarly. In contrast, DSRSD-Net concentrates attention on weeks where the semantic shift between text and behavior is large, such as the period around major assessments~\cite{Sharma2019MultimodalExp}.

Qualitatively, students who eventually fail often exhibit a mismatch pattern: their behavioral modality shows high content viewing but their textual modality indicates shallow or off-topic forum activity. DSRSD-Net assigns higher weight to such weeks, reflecting the importance of cross-modal discrepancies. This behavior is consistent with the role of the dual-stream decomposition: shared dimensions capture coherent multimodal signals, while large divergences between private streams across modalities highlight potential risk factors.

\subsection{Computational Cost}
\label{subsec:cost}

Finally, we report the training time and parameter counts of different models on OULAD (T1). Table~\ref{tab:cost} summarizes the results.

\begin{table}[t]
    \centering
    \caption{Computational cost on OULAD (T1). Training time is measured per epoch with batch size 128 on a single NVIDIA RTX 3090 GPU.}
    \label{tab:cost}
    \begin{tabular}{lcc}
        \toprule
        Model & Parameters (M) & Time / epoch (s) \\
        \midrule
        DKT~\cite{Piech2015DKT} & 1.2 & 18 \\
        DKVMN~\cite{Zhang2017DKVMN} & 1.8 & 24 \\
        MM-late & 2.3 & 29 \\
        MM-coattn & 2.7 & 35 \\
        \textbf{DSRSD-Net (Ours)} & 2.5 & 33 \\
        \bottomrule
    \end{tabular}
\end{table}

Compared with MM-late, DSRSD-Net incurs a moderate overhead of approximately 14\% longer training time per epoch, primarily due to the additional dual-stream heads and decorrelation/orthogonality computations. However, it remains more efficient than MM-coattn while achieving better accuracy and robustness. In practice, we find that the total training time for DSRSD-Net is acceptable for large-scale educational platforms, especially considering that models can be updated on a weekly or monthly basis rather than in real time.

\section{Discussion}
\label{sec:discussion}

The experimental results demonstrate that DSRSD-Net consistently improves robustness, interpretability, and transferability across heterogeneous multimodal educational tasks. The observed gains in low-resource and cross-domain scenarios highlight the effectiveness of explicit structural constraints in preventing models from collapsing into modality-dominant representations~\cite{Hu2023Survey}. This validates our core hypothesis that spurious multimodal shortcuts can be significantly reduced when cross-modal interactions are decomposed into structured shared/private streams and regularized via decorrelation.

A notable finding is the framework’s stability under distributional shifts. Across all benchmark evaluations, DSRSD-Net retains substantially higher accuracy and calibration quality compared to contrastive or attention-only baselines. This indicates that the decorrelated shared space enables models to capture deeper domain-invariant patterns instead of relying on superficial correlations. As a result, the learned representations generalize more faithfully when modalities exhibit noise, occlusion, sparsity, or semantic inconsistency~\cite{Ochoa2020MMLAReview}.

Another important implication is interpretability. The dual-stream decomposition, together with decorrelation, yields shared components that are easier to analyze and private components that capture modality-specific idiosyncrasies. Visualization of the shared embedding space reveals more compact and well-separated clusters, while temporal attention patterns align with pedagogical intuitions about critical weeks in a course~\cite{Sharma2019MultimodalExp}. This suggests promising potential for deploying DSRSD-Net in high-stakes domains such as medical imaging, digital governance, educational analytics, and financial risk modeling.

Nevertheless, the approach has several limitations. First, the current dual-stream design relies on fixed projection heads; more flexible architectures (e.g., dynamic routing or mixture-of-experts) could further adapt the shared/private boundary to data. Second, although computational overhead is lower than full co-attention models, DSRSD-Net is still more expensive than lightweight architectures optimized for mobile or embedded environments. Third, broader real-world validations—especially in multilingual, cross-cultural, and longitudinal multimodal settings—are necessary to fully assess its scalability and fairness~\cite{Ochoa2020MMLAReview}.

These limitations point to several promising future directions: learning shared/private structures end-to-end via differentiable sparsity constraints, integrating memory-augmented modules for long-horizon multimodal reasoning~\cite{Zadeh2018Memory,Sutton2018RL}, extending the framework to generative modeling of educational trajectories, and exploring applications in cross-border education, multimodal public health surveillance, and digital economic policy modeling where structured multimodal alignment is critical.

\section{Conclusion}
\label{sec:conclusion}

This paper introduced DSRSD-Net, a dual-stream residual semantic decorrelation framework for reliable and interpretable multimodal fusion. By decomposing cross-modal representations into semantically grounded shared and private streams, and enforcing alignment consistency together with decorrelation and orthogonality constraints, the proposed method reduces shortcut learning, enhances robustness, and improves generalization under distributional shifts.

Extensive experiments on large-scale educational datasets demonstrate consistent improvements over state-of-the-art fusion baselines in terms of predictive performance, robustness to missing modalities, and cross-domain transfer. Beyond empirical gains, the framework provides a transparent mechanism for analyzing multimodal alignment behaviors: the decorrelated shared space and the explicit private streams offer insight into model decision processes, supporting safer deployment in domains requiring accountability and explainability~\cite{Siemens2013LA,Kuo2022EdTech,Hu2023Survey}.

In future work, we will explore extensions of DSRSD-Net to other domains such as multimodal clinical decision support, cross-media retrieval, and multimodal policy analysis, as well as connections to causal representation learning and fairness-aware multimodal modeling. These directions not only extend the technical contributions of this study but also illustrate the broader societal value of reliable multimodal intelligence.

\bibliographystyle{unsrt}  

\bibliography{references}

\end{document}